\documentclass[manuscript,screen]{acmart}
\AtBeginDocument{%
  }

\setcopyright{acmlicensed}
\copyrightyear{2018}
\acmYear{2018}
\acmDOI{XXXXXXX.XXXXXXX}
\acmConference[Conference acronym 'XX]{Make sure to enter the correct
  conference title from your rights confirmation email}{June 03--05,
  2018}{Woodstock, NY}
\acmISBN{978-1-4503-XXXX-X/2018/06}

\usepackage{amsmath,amsfonts}
\usepackage{algorithmic}
\usepackage{algorithm}
\usepackage{array}
\usepackage[caption=false,font=normalsize,labelfont=sf,textfont=sf]{subfig}
\usepackage{textcomp}
\usepackage{threeparttable}
\usepackage{stfloats}
\usepackage{url}
\usepackage{verbatim}
\usepackage{graphicx}
\usepackage{xcolor}
\usepackage{xspace}
\usepackage{colortbl}

\usepackage{tabu}                     
\usepackage{multicol}                 
\usepackage{multirow}                 
\usepackage{float}                    
\usepackage{makecell}                 
\usepackage{booktabs}                 

\newtheorem{theorem}{Theorem}
\newtheorem{lemma}{Lemma}
\newcommand{\model}{LPS-GNN\space} 

\newcommand{\vpara}[1]{\vspace{0.04in}\noindent\textbf{#1}\xspace}

\newcommand{\hide}[1]{} 

\begin{document}

\title{\model: Deploying Graph Neural Networks on Graphs with 100-Billion Edges}

\author{Xu Cheng}
\affiliation{%
  \institution{Tsinghua University}
  \city{Beijing}
  \country{China}
}
\email{chengx19@mails.tsinghua.edu.cn}

 \author{Liang Yao}
 \authornote{The corresponding author.}
 \affiliation{
   \institution{Sun Yat-sen University}
   \city{Shenzhen}
   \country{China}}
 \email{yaoliang3@mail.sysu.edu.cn}

\author{Feng He}
 \affiliation{
   \institution{Tencent Inc}
   \city{Shenzhen}
   \country{China}}
\email{fenghe@tencent.com}

\author{Yukuo Cen}
 \affiliation{
   \institution{Zhipu AI}
   \city{Beijing}
   \country{China}}
\email{ch-zhang15@mails.tsinghua.edu.cn}

\author{Yufei He}
 \affiliation{
   \institution{National University of Singapore}
   \city{Beijing}
   \country{China}}
\email{ch-zhang15@mails.tsinghua.edu.cn}

\author{Chenhui Zhang}
 \affiliation{
   \institution{Zhipu AI}
   \city{Beijing}
   \country{China}}
\email{ch-zhang15@mails.tsinghua.edu.cn}

\author{Wenzheng Feng}
 \affiliation{
   \institution{Tsinghua University}
   \city{Beijing}
   \country{China}}
 \email{fwz17@mails.tsinghua.edu.cn}

 \author{Hongyun Cai}
 \affiliation{
   \institution{Tencent Inc}
   \city{Beijing}
   \country{China}}
 \email{laineycai@tencent.com}

 \author{Jie Tang}
 \affiliation{
   \institution{Tsinghua University}
   \city{Beijing}
   \country{China}}
 \email{jietang@tsinghua.edu.cn}

\renewcommand{\shortauthors}{Xu Cheng, et al.}

\begin{abstract}
Graph Neural Networks (GNNs) have emerged as powerful tools for various graph mining tasks, yet existing scalable solutions often struggle to balance execution efficiency with prediction accuracy.
These difficulties stem from iterative message-passing techniques, which place significant computational demands and require extensive GPU memory, particularly when dealing with the neighbor explosion issue inherent in large-scale graphs.
This paper introduces a scalable, low-cost, flexible, and efficient GNN framework called LPS-GNN, which can perform representation learning on 100 billion graphs with a single GPU in 10 hours and shows a 13.8\% improvement in User Acquisition scenarios. 
We examine existing graph partitioning methods and design a superior graph partition algorithm named LPMetis. In particular, LPMetis outperforms current state-of-the-art (SOTA) approaches on various evaluation metrics.
In addition, our paper proposes a subgraph augmentation strategy to enhance the model's predictive performance. It exhibits excellent compatibility, allowing the entire framework to accommodate various GNN algorithms.
Successfully deployed on the Tencent platform, \model has been tested on public and real-world datasets, achieving performance lifts of 8. 24\% to 13. 89\% over SOTA models in online applications. 
\end{abstract}



\received{20 February 2007}
\received[revised]{12 March 2009}
\received[accepted]{5 June 2009}

\maketitle

\section{Introduction}\label{sec:introduction}
In recent years, Graph Neural Networks (GNNs)~\cite{kipf2016semi, hamilton2017inductive, velivckovic2017graph} have drawn significant attention due to their ability to manage intricate graph-structured data. They find applications in various practical domains, including social network analysis~\cite{huang2019graph, tang2009relational}, fraud detection~\cite{dou2020enhancing, 1005055}, and recommendation systems~\cite{ying2018graph, he2020lightgcn}. However, the scalability of GNNs is a prominent challenge when dealing with extensive graphs~\cite{duan2022comprehensive}. 

Historically, designing scalable training algorithms has been a significant direction, resulting in remarkable advances. Sampling-based methods, commonly used in academic benchmarks, generally employ node-wise~\cite{hamilton2017inductive} or layer-wise sampling~\cite{fey2019fast, chen2018fastgcn}. However, node-wise sampling encounters the issue of neighbor explosion and is sensitive to hyperparameters~\cite{duan2022comprehensive}. Improving sampling efficiency requires loading the entire graph into memory, which can be difficult for billion-scale graphs. Conversely, layer-wise sampling~\cite{chen2018fastgcn} has a sparser adjacency matrix but exhibits sub-optimal performance.
PinSage~\cite{Ying_2018} faces challenges such as subgraph sampling explosion and substantial growth in cluster resources, when extends to larger graphs.

Alternatively, decoupled GNNs~\cite{wu2019simplifying,2108.10097} view feature propagation as a preprocessing step, but rely heavily o  the homophilous hypothesis, overlooking the heterogeneity and noise present in real-world graph data. In contrast, partition-based methods offer efficiency and are well-suited for parallel training and inference. Cluster-GCN~\cite{chiang2019cluster}, which depends on METIS~\cite{Karypis_1998}, struggles with super large graphs, and the stochastic multiple partitions scheme requires the adjacency matrix to be stored in memory, necessitating the support of a distributed system. 
Distributed systems, such as PyTorch-BigGraph~\cite{lerer2019pytorch}, AliGraph~\cite{zhu2019aligraph}, DistDGLv2~\cite{2112.15345}, and Angel Graph~\cite{jiang2020psgraph}, all employ partition-based data parallelism. 
However, a comprehensive examination and analysis of large-scale graph partitioning algorithms designed for GNNs is absent.

Existing graph partition algorithms,  including spectral-based~\cite{von2007tutorial, DBLP:journals/tkde/FengLKC24}, flow-based~\cite{yu2010scalable}, and recursive bisection algorithms~\cite{Karypis_1998}, strive to optimize three key objectives: execution speed, minimum cut, and partition balance. However, the computational time required for large graphs increases significantly. Moreover, obtaining globkal information necessary for achieving partition balance becomes challenging, and significant disparities in the distribution of local subgraphs further complicate the attainment of a minimum cut. METIS utilizes the multilevel graph bisection technique for graph partitioning and achieves SOTA results on all three metrics for smaller graphs (10 million nodes). ParMetis~\cite{karypis1998parallel}, an optimized parallel version of METIS, can handle graphs with up to 240 million nodes~\cite{2112.15345}. SpecGDS~\cite{DBLP:journals/tkde/FengLKC24} extended its ability to handle 1.47 billion edges. Nonetheless, these capabilities still fail to handle graphs of 100 billion edges. 
We also examine community detection algorithms like the Label Propagation Algorithm (LPA), which can handle large-scale graphs, but do not guarantee balanced community sizes. 

To address these shortcomings, this paper introduces a new computing framework, LPS-GNN~\footnote{Our source code is available at ~\url{https://github.com/yao8839836/LPS-GNN}}, to render GNNs scalable to super-large graphs, which can train a graph with 100 billion edges on a single P40 GPU (24GB). 
It comprises three components: partitioning, sub-graph augmentation, and GNNs. The partitioning component (LPMetis) integrates the computational speed of the LPA algorithm and the partition balance of METIS through a Multi-level framework. Considering the redundancy in graph information, LPMetis prioritizes execution speed and partition balance, tolerating a certain degree of edge cutting. In the graph augmentation component, a hypergraph representation is used to capture the global information of large graphs, while subgraph augmentation complements critical local information to enhance the performance of GNNs and mitigate information loss caused by edge cutting. Finally, through empirical experiments, we demonstrate that partitioning algorithms based on information propagation can better adapt to GNNs. We also validate the effectiveness of the \model framework in various industrial scenarios of different scales.
The following summarizes the contributions of this work:
\begin{itemize}
\item LPMetis is a distributed and parallel heuristic suited for graph partition and achieves superior tradeoffs in terms of partition balance, minimum cuts, and training speed, enabling GNNs to process 100-billion-edge graphs. 
\item \model improves the efficacy of GNNs by sub-graph sampling and enhancing sub-graphs through structural refinement and feature augmentation, which avoids traversing entire graphs and enhances performance. 
\item \model shows outstanding performance in both supervised and unsupervised tasks in three industrial scenarios. 
Empirical results demonstrate that the proposed algorithm enhances baseline models by up to 8.24\% in conversion rate for friends recommendation, as well as 11.94\% in precision for cheating users detection and up to 13.89\% in precision for ad user acquisition.
\end{itemize}

\section{RELATED WORK}
\subsection{Graph Neural Networks}
GNN works can be categorized into three groups based on their approaches to handling large-scale graphs.
The first category consists of advanced system architectures tailored specifically for large-scale graphs. 
This includes PaGraph~\cite{lin2020pagraph}, Betty~\cite{yang2023betty} and ByteGNN~\cite{zheng2022bytegnn}. 
These systems utilize data parallelism, multi-GPU, and parameter servers. However, large-scale graphs must be partitioned into smaller subgraphs to employ these systems. 
To our knowledge, our study is the first to tackle the partitioning of 100-billion graph networks.
Others employ sophisticated system design and sampling algorithms.
MariusGNN~\cite{waleffe2023mariusgnn} introduces a novel data structure called DENSE and proposes a transmission sampling method to improve the computational efficiency of GNNs.
Ginex~\cite{Park_2022} utilizes the Belady algorithm to cache features in memory and employs a sampling algorithm to enhance efficiency, with a focus on improving data transfer between SSD and memory.
The second category decouples feature propagation from nonlinear transformation operations, with examples including SGC~\cite{wu2019simplifying}, APPNP~\cite{klicpera2018predict}, and GAMLP~\cite{zhang2022graph} and FS-GNN~\cite{maurya2022simplifying}. Although these methods exhibit promising performance on academic graph benchmarks~\cite{hu2020open}, their effectiveness may be constrained on more intricate and realistic graphs, primarily due to their pronounced homophily.
The third category leverages sampling, parameter sharing, and partitioning to balance training efficiency and effectiveness. Sampling techniques employed by GraphSAGE~\cite{hamilton2017inductive}, PinSAGE~\cite{Ying_2018}, FastGCN~\cite{chen2018fastgcn}, GraphSAINT~\cite{zeng2019graphsaint} and BNS-GCN~\cite{wan2022bnsgcn} pose various challenges. For instance, METIS imposes scalability limitations of BNS-GNN, the neighbor explosion issue leads to exponential growth in computation time and memory usage with an increasing depth of the GNN. Furthermore, sampling-based methods require the use of MapReduce to prevent duplicate computations~\cite{Ying_2018}. Partitioning techniques, such as METIS~\cite{karypis1998fast} used by Cluster-GCN~\cite{chiang2019cluster}, aim to accelerate GNN training, but the between-cluster link sampling~\cite{chiang2019cluster} restricts their applicability to super-large graphs. 
GCN-LPA~\cite{wang_unifying_2020} strives to unify GCN and LPA, interpreting GCN as feature propagation and LPA as label propagation.
Incorporate the loss of LPA as a regularization term in the GCN loss function to jointly optimize the overall objective. Recent studies like LLM-GNN~\cite{chenlabel} incorporate powerful language understanding abilities of Large Language Models (LLM) into GNN learning.
\subsection{Graph Partition Methods} 
Graph partitioning, a pivotal issue in computer science and graph theory, involves segmenting a graph into clusters or partitions.
Graph partition algorithms evolving over time with varying optimization focuses, can be classified into three types. 
Numerous edge-cut partitioning algorithms, such as METIS~\cite{karypis1998fast}, Fast-unfolding~\cite{blondel2008fast}, Fast-unfolding with density~\cite{chen2018network}, and Label Propagation~\cite{raghavan2007near}, have been introduced. LPNMF~\cite{9044402} based on Nonnegative Matrix Factorization. Nevertheless, these algorithms possess inherent limitations that hinder load balancing on large-scale graphs or minimum cut guarantees. 
With the surge in graph data, executing computations on a single machine becomes increasingly infeasible due to high memory demands~\cite{adoni2020survey}. 

The second category comprises emerging scalable graph partition algorithms, aimed at improving large graph processing efficiency through parallel or distributed computing, typically handling datasets of millions. Examples include \textit{Guo et al.}~\cite{10137050} integrating graph partitioning with multi-GPU parallel computing, ParMetis utilizing the OpenMPI framework~\cite{karypis1998parallel}, Recursive Graph Partitioning (RGP)~\cite{lin2021large} implemented in Spark using iterative METIS, and LShape Partitioning~\cite{9186333} employing MapReduce for Skyline Query Processing. Other approaches feature pipeline architecture~\cite{chen2018g} and application-specific decoupled graph partitioning techniques~\cite{10137050}.

The third type integrates GNN-compatible partition algorithms for specific scenarios. Examples include PaGraph~\cite{lin2020pagraph}, which enhances training scalability in multi-GPU environments, and ROC~\cite{jia2020improving}, which uses an online linear regression model for load balancing. Algorithms like Betty~\cite{yang2023betty}, BGL~\cite{285052}, and ByteGNN~\cite{zheng2022bytegnn} cater to specific applications from micro-batch optimization to sampling load balancing. FMVPG~\cite{9428598} constructs a prototype graph for spectral rotation, and Graph Embeddings~\cite{9169850} serve as partitions.
NeuGraph~\cite{ma2019neugraph} proposes a multi-GPU GNN framework, transforming GNNs into dataflow graphs for large-scale computations.

\begin{figure}[!t] %
\centering
\includegraphics[width=0.73\linewidth]{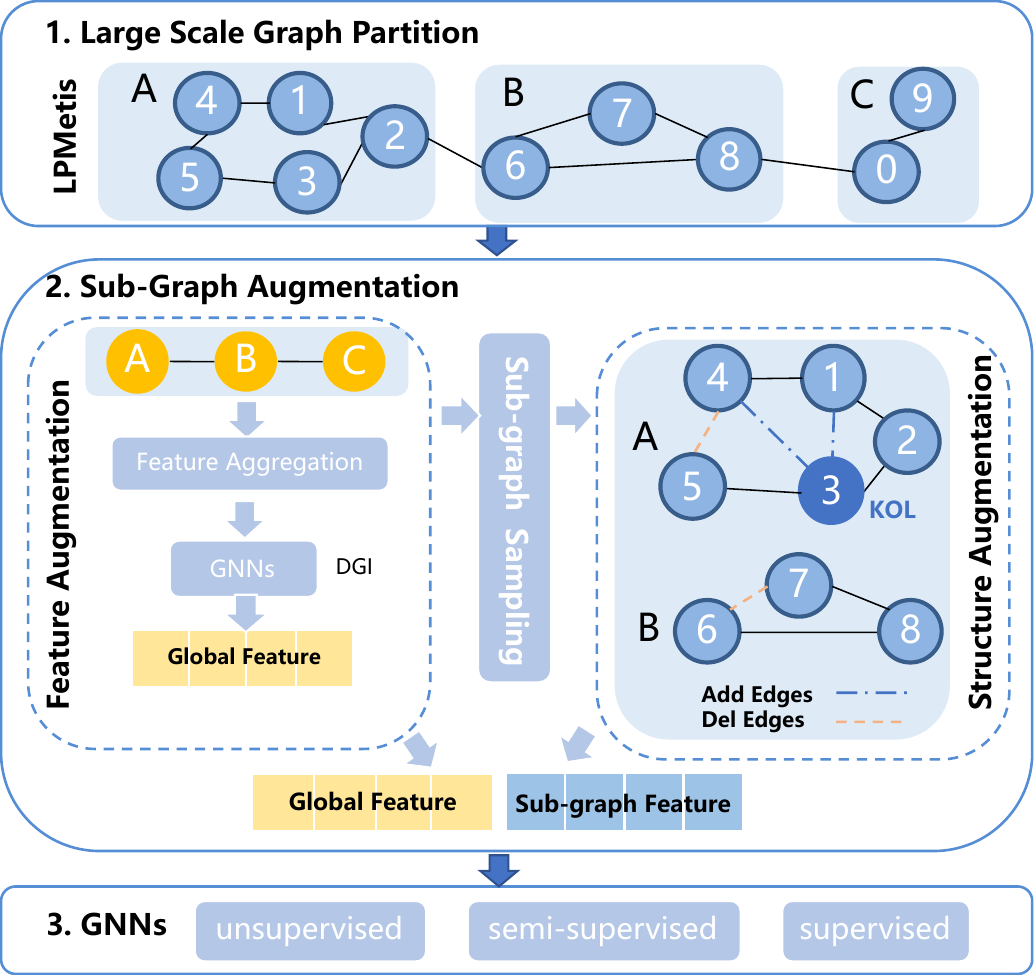} 
\caption{Overview of \model framework. 1) LPMetis achieves SOTA results at load balancing, minimum cuts, and efficiency. 2) Global feature augmentation and subgraph structure refinement improve the performance of GNNs. 3) GNNs are selected based on the tasks.}
\label{fg:state}
\end{figure} %
\section{ Preliminaries}
\vpara{Notations.}
Let $G= (V, E)$ be a graph, which consists of $ N = |V|$ vertices and $|E|$ edges. The weight between nodes $v_i$ and $v_j$ is denoted as $e_{ij}$. The sparse matrix $A \in \mathbb{R}^{N\times N}$ represents the adjacency matrix. The feature matrix $X \in \mathbb{R}^{N\times F} $ contains an $F$-dimensional feature vector. 
$\mathcal{N} (v_i)$ represents the neighbors of node $v_i$, $N_{out} (v_i)$ and $N_{in} (v_i)$ represent the out-degree and the in-degree of node $v_i$. 
$C_{v_i}$ signifies the community label of node $v_i$, indicating its affiliation within the potentially complex communal structure inherent to $G$.
A large-scale graph $G$ is divided into $K$ disjoint subgraphs $P(G) = \{g_1, g_2, g_3, \ldots, g_K\}$, where $|v_{g_k}|$ denotes the number of nodes in the subgraph $g_k$. 
In the context of graph coarsening, $P(G)$ is used as input to generate a coarsened graph $\tilde{G}_k = (\tilde{V}_k, \tilde{E}_k)$, where 
$|\tilde{V}_k|$ reflects the aggregated quantity of nodes from the corresponding subgraph $g_k$ in $P(G)$. 
Moreover, $\tilde{E}_{k_1 k_2}$ represents the cumulative weight of edges that connect the different subgraphs $g_{k_1}$ and $g_{k_2}$ within $P(G)$.
${\tilde{v}_{k}}$ and ${\tilde{e}_{k}}$ represents the nodes and edge in subgraph $g_k$.
$W_l$ is the learnable parameters of GNN at the l-th layer.
%

METIS is a multilevel graph bisection technique with three key phases: coarsening, partitioning, and uncoarsening. 
Coarsening transforms the original graph $G$ into a series of smaller graphs $ {\tilde{G_0}, \tilde{G_1}, \ldots, \tilde{G_m}} $, where $ |\tilde{V_0}| > |\tilde{V_1}| > \ldots > |\tilde{V_m}| $. Edge-based coarsening (Algorithm~\ref{alg:PBGNN}, Line 2; Algorithm~\ref{alg:LPMetis}, Line 4) focuses on edge weights, while node-based coarsening~\cite{karypis1998fast} (Algorithm~\ref{alg:LPMetis}, Line 6) balances node counts across subgraphs (see Section \ref{secDifferentTypesofCoarsening}). Partitioning splits the final coarsened graph $\tilde{G_m} $ into two subgraphs, and uncoarsening projects the partition back to $ \tilde{G_0} $.
\begin{figure*}[!t]
  \centering
  \includegraphics[width=0.87\linewidth]{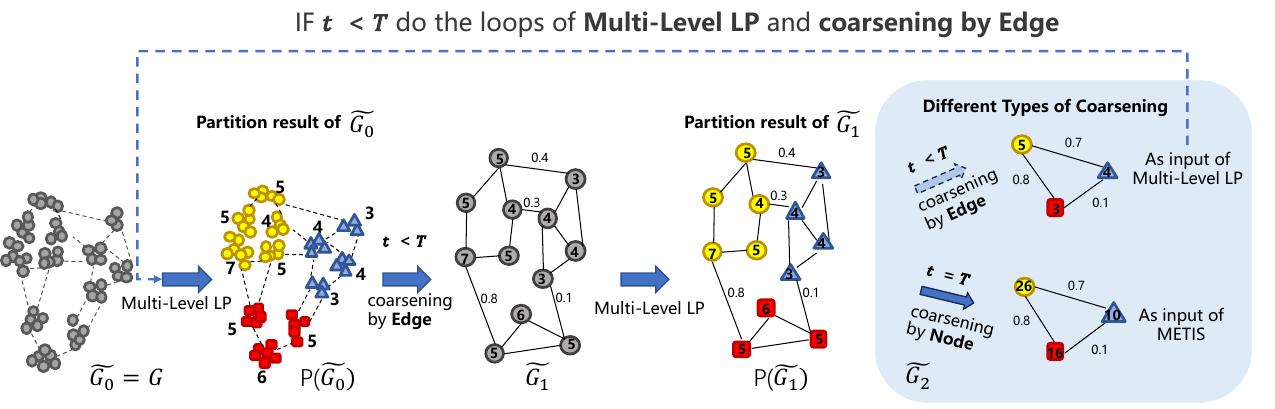} 
  \caption{The process flow of LPMetis algorithm. Initially, $\tilde{G_0}$ is utilized as input for the Multi-Level LP algorithm, resulting in $P(\tilde{G_0})$. Next, $\tilde{G_1}$ is constructed through the "coarsening by edge" approach and serves as input for Multi-Level LP. This iterative procedure continues until t=T. Then, $P(\tilde{G_2})$ is generated using the "coarsening by node" strategy. The section titled "Different Types of Coarsening" outlines the differences between the two coarsening methods when the input is $P(\tilde{G_1})$. } 
  \label{fig:lpmetis}
\end{figure*}
%
%
%

The Label Propagation Algorithm (LPA), a semi-supervised method, assigns labels to unlabeled data via a parallelizable node update mechanism, enhancing performance. However, LPA may exhibit oscillation and non-convergence in bipartite networks ($\chi(G) = 2$) (Figure ~\ref{fig:ss} (b)).
The green node updates to yellow due to yellow neighbors, causing all yellow nodes to switch to green, resulting in oscillation and hindering convergence. While asynchronous updates mitigate this, they compromise accuracy, prolong convergence, and reduce efficiency for large-scale graphs.
To address this, we propose the Multi-Level Label Propagation algorithm (Section ~\ref{secLargeScaleGraphPartition}).
\section{METHOD}
The \model framework is guided by a unified perspective on semi-supervised learning~\cite{agovic2009unified}, which suggests that label propagation, graph-cuts, and manifold embeddings can be seen as specific instances of generalized label propagation under certain constraints.
%
%
\subsection{\model Overview}
We developed \model for large-scale graph computations in homogeneous networks (Figure~\ref{fg:state}).
Leveraging Tencent's distributed computing infrastructure, \model utilizes platforms for GNN preprocessing and graph partition, including a scheduling platform capable of executing SPARK tasks and a distributed file system. Then, training is performed on a single GPU machine. 
Algorithm~\ref{alg:PBGNN} illustrates that LPMetis-derived subgraphs for graph convolution resemble those sampled via random walks, similar to PinSage~\cite{Ying_2018}. A global coarsened graph is built from LPMetis partitions. Line 3 employs unsupervised/self-supervised GNNs (e.g., DGI~\cite{velickovic2019deep}) for node embeddings with global position information. Lines 4-9 involve subgraph augmentation and GNN training using sampled partitions per batch.
\begin{algorithm}[!t]
    \caption{ Partition-based Billion-scale Graph Neural Networks (\model ) }
    \label{alg:PBGNN}
    \begin{algorithmic}[1]
        \REQUIRE {Graph $G= (V, E)$, feature $X$, label $Y, K, M $}
        \ENSURE  { Node representation $\bar{X}$}
        \STATE  Cut graph nodes into K partition $g_1, g_2, \ldots, g_{K}$ by LPMetis;
        \STATE Construct $\tilde{G} = \operatorname{Coarsen} (\{g_1, g_2, \ldots, g_{K}\}, ``Edge", G)  $
        \STATE Global embedding $X^{global} = GNN_{unsup} ( X_{\tilde{G}}, A_{\tilde{G}} )$ 
        \REPEAT
            \STATE Randomly sample $ g_m$ from $ \{g_1, g_2, \ldots, g_{K}\}$
            \STATE $\hat{X}_{g_m} \gets [X_{g_m}; X^{global}_{g_m} ]$
            \STATE  $ \hat{A}_{g_m}  \gets  \operatorname{StructAugmentation} (  A_{g_m}  )$ \ \ based\ on\ Section \ref{sec:StructAug}
            \STATE  $ { \{W_l\} }^L_{l=1} \gets GNN_{super} (  \hat{X}_{g_m}, \hat{A}_{g_m}, Y_{g_m}, \ { \{W_l\} }^L_{l=1} )$
        \UNTIL{Convergence}
        \ENSURE  {Return $ { \{W_l\} }^L_{l=1}$}
    \end{algorithmic}
\end{algorithm}
%
%
%
\begin{algorithm}[!t]
    \caption{ Label Propagation with Metis (LPMetis) }
    \label{alg:LPMetis}
    \begin{algorithmic}[1]
        \REQUIRE {$G, K, T$}

        \ENSURE  {Cut graph nodes into $K$ partitions $g_1, g_2, \ldots, g_{K}$}
 
        \STATE  $\tilde{G} \gets G $ 
        \FOR {$t=0$ to $T$}
        
        \STATE  $ \{g_1, g_2, \ldots, g_{M}\} \gets MultiLevelLP ( \tilde{G}, T) $

        \STATE   $ \tilde{G} \gets  \operatorname{Coarsen} ( \{g_1, g_2, \ldots, g_{M}\}, "Edge", \tilde{G}) $ 
        \ENDFOR
        \STATE   $ \tilde{G} \gets  \operatorname{Coarsen} ( \{g_1, g_2, \ldots, g_{M}\}, "Node", \tilde{G}) $ 
        
       \STATE  $ \{g_1, g_2, \ldots, g_{K}\} \gets \operatorname{METIS} ( \tilde{G}, K) $
       \STATE  Mapping the subgraph ID $K$ back to node $v$ in graph $G$
       \ENSURE  {$K$ partitions $ \{g_1, g_2, \ldots, g_{K}\} $}
    \end{algorithmic}
\end{algorithm}
\subsection{Algorithms}
\subsubsection{Label Propagation with METIS (LPMetis)} \label{secLargeScaleGraphPartition}
LPA, known for its fast speed, can be used as a graph-cuts algorithm and is better aligned with the message passing mechanism of GNNs~\cite{wang_unifying_2020}, making it a promising choice (Table~\ref{tab:OGB-PARTITION2}). 
Additionally, Metis and RGP have demonstrated the effectiveness of the multi-level approach. Therefore, by combining the strengths of LPA and Metis through the multi-level technique, and incorporating the partition-based framework, we can efficiently handle large-scale data cheaply.
METIS consistently applies the same coarsening (maximal matching) and partitioning methods across all levels, while LPMetis (Figure~\ref{fig:lpmetis}) employs diverse operators to address specific challenges, such as Multi-Level Label Propagation and METIS. The Multi-Level Label Propagation operator facilitates rapid segmentation of large graphs (Algorithm \ref{alg:LPMetis}, Lines 1-5), and the METIS operator ensures balanced node distribution among subgraphs (Lines 6-7). Additionally, LPMetis directly maps subgraph IDs back to the node set $\mathcal{V}$, eliminating the need for further refined or projected partitioning steps (Line 8). The number of subgraphs $K$ is typically determined heuristically, based on single GPU memory capacity, graph edge scale, and desired nodes per subgraph.
\subsubsection{Multi-Level Label Propagation} 
\begin{figure}[!t]
  \centering
  \includegraphics[width=0.87\linewidth]{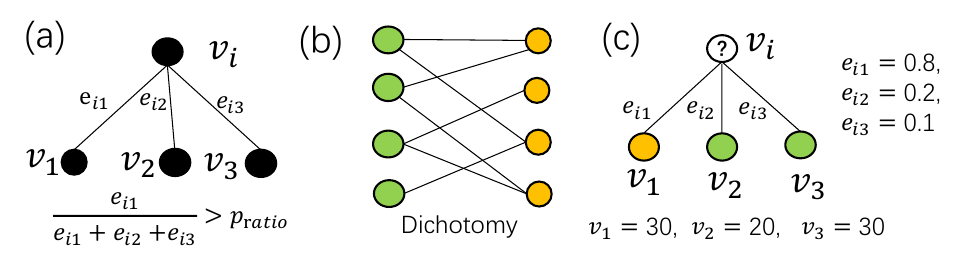} 
  \caption{The key steps and concepts in Multi-Level LP}
  \label{fig:ss}
\end{figure}

Multi-Level LP aims to preserve more edges during partitioning, prevent dichotomous scenarios, and accelerate graph coarsening (Algorithm \ref{alg:CLP}). We initialize node labels $C_v$ with their respective IDs (Line 1). Given the graph's inherent redundancy and noise, we establish a threshold for edge retention or removal (Line 4). The relative weight $P_{ij}$ of the edge $e_{ij}$ is calculated using Eq. (\ref{equ:p}), and edges with $P_{ij}$ exceeding $P_{ratio}$ are retained (Figure~\ref{fig:ss}(a)). 
Removing all edges with weights less than $P_{ratio}$ degrades GNN performance. Thus, we randomly retain edges where $P_{rand} < P_{b}$, with $P_{rand}$ sampled from a uniform distribution between 0 and 1.
Both $P_{\text{ratio}}$ and $P_{\text{b}}$ are hyperparameters, typically set heuristically, exhibiting insensitivity within their contexts. For unweighted graphs, the edge weight is set to 1.
\begin{equation}
\label{equ:p} 
P_{ij} =  
    \begin{cases} 
    \frac{e_{ij}}{\sum_{m \in \mathcal{N} (v_i)} e_{im}} & \text{ if } P_{ij}  \geq P_{ratio} \  or \ P_{rand} < P_{b} \\
    0  & \text { otherwise } 
    \end{cases}
\end{equation}

In a dichotomous scenario (Figure~\ref{fig:ss} (b)), LPA selects the most frequent label $C$ among neighboring nodes, disregarding edge weights and subgraph size. For example, in Figure~\ref{fig:ss} (c), node $v_i$ has three neighbors with green and yellow labels; LPA favors the green label due to its higher frequency (two out of three neighbors). Incorporating edge weights in label selection prevents dichotomous situations (Figure~\ref{fig:ss} (b)), adapts to weighted graphs, and reduces cumulative errors during coarsening. Multi-Level LP synchronously updates edge weight $e_{ij}$ and node value $v_i$ (Line 3), as illustrated in Eq.~(\ref{equ:vlabel}) (omitting explicit iteration indices t and t-1 for simplicity). The function $f$ aggregates weighted label values, returning the yellow label with the highest score.
\begin{equation}
\label{equ:vlabel} 
C_{v_i} = f\left ( \frac{e_{i1}}{v_{i1}} C_{v_{i1}}, \ldots, \frac{e_{im}}{v_{im}} C_{v_{im}}, \ldots, \frac{e_{in}}{v_{in}} C_{v_{in}}  \right) \  
\end{equation}
%

For winstance, if $C_{v_{i2}}$ and $C_{v_{i3}}$ are both green, the weight coefficients $\frac{e_{i2}}{v_{i2}} = 0.01$ and $\frac{e_{i3}}{v_{i3}} = 0.003$ are combined to yield $0.013$ as the new weight coefficient for the green label. 
Ultimately, $f$ selects the yellow label due to its higher weighted average value of $\frac{e_{i1}}{v_{i1}} = 0.026$, compared to the green label's $0.013$.
The default value of $v_i$ is $1$. In the case of a coarsened graph $\tilde{G_t}$, $v_i$ denotes the number of nodes in the partition of $\tilde{G_{t-1}}$ (i.e., $v_1 =30$).
\begin{algorithm}[!t]
    \caption{ Multi-Level Label Propagation (Multi-Level LP) }
    \label{alg:CLP}
    \begin{algorithmic}[1]
        \REQUIRE {$G= (V,E), T$}
        \ENSURE  {Cut graph nodes into $M$ partitions $g_1, g_2, \ldots, g_{M}$}
        \STATE Initialize the labels at all nodes in the network. For a given node $v$, $C_v \gets v $  
        \FOR {$t=0$ to $T$}
            \STATE Update each node's label with the highest frequency label among neighbors based on Eq. \ref{equ:vlabel}
            \STATE Spread the labels along the edges. Delete edge $ e_{ij} $ if $ P_{ij}=0$ based\ on\ Eq. \ref{equ:p}
        \ENDFOR
    \STATE Assign nodes $V$ into $M$ clusters $ g_1, g_2, \ldots, g_{M}$ according to $C_{v}$.  
    \ENSURE  {$M$ partitions \{$g_1, g_2, \ldots, g_{M}$\}}
    \end{algorithmic}
\end{algorithm}
%
\subsection{Graph Augmentation}
To enhance the performance of GNNs, in addition to preserving more edges like LPMetis, we also combine features and structure of graph to boost the GNN's performance.

\subsubsection{Feature Augmentation}
The output of LPMetis produces a coarsened graph denoted as $\tilde{G}$. 
Each node in $\tilde{G}$ represents the position and features of a subgraph $g$ in the original graph.
For instance, in Figure~\ref{fg:state}, the global coarsened graph consists of three nodes: A, B, and C.
Node A aggregates the features of all nodes in the corresponding subgraph.
The weight of an edge $\tilde{E}_{A B}$ between $g_A$ and $g_B$ is determined by summing all the edges connecting the subgraphs, represented as $\sum e_{ij}$($i\in v_A$ and $j\in v_B$).
Subsequently, the unsupervised algorithm DGI~\cite{velickovic2019deep} is employed to extract the global information $X^{global}$ from the coarsened graph $\tilde{G}$. Other types of unsupervised GNNs are also suitable. 
\subsubsection{Structure  Refinement} 
\label{sec:StructAug}
PageRank assesses a node's influence on its neighbors; the features of the bottom $5\%$ nodes with the least influence diminish in significance during propagation and offer minimal utility in predicting other nodes, thereby justifying their removal to improve performance.
We run PageRank~\cite{bahmani2010fast} on the subgraph as $ \operatorname{PR}\left (v_i\right)=\frac{1- \alpha }{|V|}+ \alpha \sum_{v_j \in \mathcal{N}\left (v_i\right)} \frac{\operatorname{PR}\left (v_j\right)}{ N_{out} \left (v_j\right)}$ 
and delete node $v_{i}$, if $PR (v_i)$ falls in the top $5\%$ smallest ($\alpha$ is a residual probability, set to 0.85). 
We also experimented with node2vec~\cite{grover2016node2vec}, but the results were not as effective as PageRank.
\subsection{Training}
\subsubsection{LPMetis}
By employing the distributed LPMetis implemented in Scala on a Spark cluster, subgraphs of similar size were then stored in a distributed file system together with the global coarsened graph. Initially, we planned to distribute the divided subgraphs among different GPUs and enable parameter sharing through a parameter server for distributed training. However, by chance, we realized that training could be accomplished by directly sampling the subgraphs and using only one P40 GPU. And convergence can be attained with a mere 5\%-10\% of the total number of subgraphs. Thus, we chose uniform sampling and avoided exploring other sampling methods due to their increased computational complexity. The sampled subgraphs underwent subgraph augmentation, whereby the global coarsened graph node embeddings were concatenated with each subgraph node's features. Each individual subgraph constituted a distinctive batch, serving as input for GNNs. The subsequent process mirrored other GNN algorithms, with the selected GNN algorithm retaining a consistent loss function. During inference, predictions were made sequentially for all subgraphs.
\subsubsection{GNN Phase}
The choice of the optimal GNN algorithm relies on the nature of the downstream tasks. 
In this paper, we primarily test unsupervised tasks. 
Our framework, \model, can easily support both the Deep Graph Infomax (DGI)~\cite{velickovic2019deep} (used by default) and the recently proposed GraphMAE~\cite{hou2022graphmae}.
Supervised GNNs can also be employed to extract global features from large graphs, but one should consider the problem of uneven label distribution across subgraphs, which is not discussed in this paper. 
We tested supervised GNNs in online experiments, achieving satisfactory results.
Empirically, node classification benefits from balanced subgraph labels, while link prediction benefits from denser subgraphs.
It should be noted that regardless of the downstream task, we recommend using unsupervised Graph Neural Networks (GNNs) during the Feature Aggregation stage to extract the global features of the large graph. 
\subsection{ Theorem Analysis}
The convergence of \model is intimately related to the mechanism of message propagation, the properties of adjacency matrices, and the configuration of the GNN model. 
To simplify the analysis, we make the following assumptions:
a) The activation function $\sigma$ is nonlinear and possesses Lipschitz continuity with $\| L \| \leq 1$. 
b)  $\tilde{A} = D^{-1/2} (A + I) D^{-1/2}$ denotes the renormalized adjacency matrix of $G$. The eigenvalues of $\tilde{A}$ lie within the range $0 < \lambda_i \leq 1$, and its spectral norm satisfies $|\tilde{A}| \leq 1$.
c) The GNN algorithm used in LPS-GNN guarantees convergence, and the parameter $\|W\| \leq 1$ is maintained through weight regularization or structural design (e.g., residual connections). The weight change between layers is $\|W^{(k+1)} - W^{(k)}\| \leq \epsilon$, and as the iterations converge, $\epsilon \ll 1$.
\begin{theorem} \label{theorem:1}
 Let the propagation formula for GNNs be given by $H^{(k+1)} = \sigma(\tilde{A} H^{(k)} W)$, where $H_{g_i}$ is hidden layer matrix of subgraph $g_i$. Then, 
$$
\|H_{g_i}^{(k+1)} - H_{g_i}^{(k)}\| < \|H_{g_i}^{(k)} - H_{g_i}^{(k-1)}\|,
$$
\model is proven to convergxe
\end{theorem}
\subsection{Time and Space Complexity}
\subsubsection{LPMetis}
The time complexity of the Multi-Level/ Label Propagation (Multi-Level LP) algorithm in each iteration is $O(|V| + |E|)$~\cite{raghavan2007near}, which is the same as the LPA. However, empirical evidence suggests that Multi-Level LP requires only 2 or 3 iterations to converge, whereas LP requires more than 5 iterations~\cite{raghavan2007near}. 
After the last iteration, Multi-Level LP transforms the subgraphs into a coarsened graph $\tilde{G}$ with thousands of nodes and passes it as input to METIS. The time complexity of this phase can be treated as constant ($O(1)$) and is negligible. The final mapping step has a $O(|V|)$ time complexity. 
The overall time complexity of LPMetis is $O(|V|+|E|)$, which is similar to that of LPA. However, LPMetis outperforms LPA by converging faster and performing iterations on successively coarser graphs. 
Furthermore, LPMetis demonstrates a nearly linear time complexity, detailed insights of which can be found in Section~\ref{secEXPPartitionOnGraphPartition}.
\subsubsection{\model}
In subgraph augmentation, feature augmentation processes a significantly smaller coarsened graph within 10 thousand nodes and only requires embedding computation once during the training step, resulting in a constant time complexity. On the other hand, structure refinement has a time complexity of $O(\lambda |E_k|)$ based on PageRank, where $\lambda$ is the number of iterations and $|E_k|$ is the number of edges in subgraph $g_k$. Only a small number of sampled subgraphs require structure refinement. Hence, the time spent on structure refinement is much less significant compared to the entire algorithm. For simplicity, we neglect the time cost of feature augmentation and structure refinement, making \model equivalent to the partition-based GNN framework such as Cluster-GCN~\cite{chiang2019cluster} costing $O(L|E|F + L|V|F^2)$ per epoch. 
Regarding space complexity (Table ~\ref{tab:LS-GNN SOTA}), \model loads $b$ samples, which incurs a memory cost of $O(bLF + LF^2)$.
\section{Experiments} \label{secExperiments}
In this section, we present extensive experiments conducted for the proposed LPS-GNN, along with its core module LPMetis.
The experiments aim to address the following three questions:
\begin{itemize}
\item {\verb|Q1|}: How does LPMetis perform compared to other SOTA graph partitioning approaches? 
\item{\verb|Q2|}: What is the contribution of each module within the \model framework? 
\item{\verb|Q3|}: How does \model compare to the SOTA GNN approach on large-scale, real-world graphs? 
\end{itemize}
\subsection{Experimental Setup} \label{secExpSetup}
\subsubsection{Dataset} \label{secExpData}
To assess the efficacy and efficiency of LPS-GNN, we initially tested the performance of LPMetis using ogbn products~\cite{hu2020open}, which was downloaded from the OGBN \footnote{\url{http://manikvarma.org/downloads/XC/XMLRepository.html}}. This dataset\footnote{\url{https://ogb.stanford.edu/docs/nodeprop/}} comprises a co-purchasing network of Amazon products and facilitates comparison with different versions of Cluster-GCN. 
However, larger public datasets may not meet the requirements of our business scenarios. So, we construct several industrial datasets of varying sizes. Dataset $A_i$ contains hundreds of millions of edges, while dataset $B_j$ ranges from billions to tens of billions of edges. 
The subsequent experiments primarily involve four datasets in Table \ref{tab:data}: ogbn-products, $A_4$, $B_2$, and $C$, which are marked in gray.

$A$ and $B$ are sampled from eight different games. Some networks are densely connected, while others are relatively sparse. This diversity is advantageous for evaluating the algorithm's robustness. 
To evaluate LPS-GNN's scalability, we sampled users from 30+ games, linked their social networks via a unified account system, creating a super-large graph, Class $C$, with hundreds of billions of edges.
The node features in Tencent games consist of 188 dimensions and primarily capture information related to gaming activities and payment data.
Edge weights measure player intimacy in a game based on interactions like joint sessions, gift exchanges, and chatting. Each game has unique criteria for this measurement, normalized to a [0, 1] scale.
%
\begin{table}[!th]
  \caption{ Statistics of the dataset }
  \centering
  \label{tab:data}
  \begin{tabular}{crr}
    \toprule
    Dataset   & \#Nodes & \#Edges  \\
    \midrule
\rowcolor{gray!50} \cellcolor{white} 
ogbn-products & 2,449,029 & 61,859,140 \\
    $A_1$  & 5,003,843 & 18,781,099 \\
    $A_2$  & 31,968,702 & 83,407,182 \\
    $A_3$  & 11,854,302 & 114,243,430 \\
\rowcolor{gray!50} \cellcolor{white}$A_4$  & 46,299,769	 & 791,779,286 \\
    $A_5$  & 52,074,681  &  879,904,294 \\
    $B_1$   & 156,507,292   & 5,948,158,120   \\
\rowcolor{gray!50} \cellcolor{white}$B_2$   & 323,698,072   & 19,992,344,560  \\
    $B_3$   & 523,602,446   & 57,576,249,243    \\ 
\rowcolor{gray!50} \cellcolor{white}$C$   & 866,799,588  &  114,194,237,292   \\ 
  \bottomrule
\end{tabular}
\end{table}
\subsubsection{Computation Resources}
We utilized a Parameter Server (PS) with 10 workers and a Spark cluster with 70 executors to facilitate the experiments. Each Spark executor has 2 CPU resources and 30GB of memory, while each PS worker utilizes 2 CPUs and 6GB of memory. GNNs were trained on a Linux server featuring 253GB of RAM, a P40 GPU, and a 48-core AMD CPU. Metis runs on a Linux server, utilizing 1 CPU and 100G of memory.
Partition experiments for super-scale graph $ C$ were conducted on a Spark cluster with 100 executors, each having 2 cores, 40GB memory, and 40GB driver memory.
\subsubsection{Evaluation Metrics and Parameter Settings} \label{secEXPMetricsSetting}
We utilize four edge-cut metrics such as Balancing (BAL), Minimum Edge-Cuts (EC), Runtime and Standard Deviation (STD) to gauge the efficiency and effectiveness of edge-cut partitioning.
$BAL = \text { MaxLoad } / \frac{|E|}{k}$,  where MaxLoad denotes the number of edges of the maximum partition~\cite{10.1145/2556195.2556213}.
$EC = (\# \text { Edges Cut by Partition} ) /\ (\# \text { Total Edges} )$. As EC increases, BAL (edge-based metric) decreases rapidly, implying that BAL comparisons are valid only when EC values are comparable.
$STD = \sqrt{  \sum{(|V_i| - \mu)}^2 / (k-1) },\ \mu = \frac{|V|}{k} $ is node-based and independent of EC.
STD resembles another BAL metric $( (1 \pm \epsilon)\frac{|V|}{k})$ ~\cite{188418}, but it is more succinct and interpretable. 

The LPMetis uses four hyperparameters: $P_{ratio}=0.5$, $P_{bound}=0.1$, $T=2$, and $K=6000$. These parameters are generally not sensitive and can largely remain unaltered, with the exception that K may need adjustment based on the size of the dataset. 
Notably, subgraphs with fewer than 30,000 nodes may cause GNN instability.
The performance of \model is evaluated using metrics such as AUC, Precision (P), Recall (R), and F1. The embedding dimension for all models is fixed at 128, and the batch size corresponds to the sub-graph node size. The number of GNN layers is set at 4, selected from a range between 3 and 7. For training, we use the Adam optimizer with a learning rate of 0.001 and a decay rate of 0.
For DGI, we have 2 MP layers with 32 dimensions in each MP layer, trained for 3 epochs.
\subsection{Experiment on Graph Partition} 
\label{secEXPPartitionOnGraphPartition}
We first evaluated the performance of LPMetis using the publicly dataset ogbn-products and then expanded the testing to 9 datasets with a range from 18M to 114B. 
\begin{table}[!t]
\centering
\caption{The performance of LPMetis on ogbn-products }
\label{tab:OGB-PARTITION2}
\begin{tabular}{lcc}
\toprule
\textbf{Model}   & \textbf{Val acc} & \textbf{Test acc}  \\ 
\midrule
LR                & 0.51221  &   0.40692       \\ 
GraphSAGE         & 0.84668  &   0.66330       \\ 
LLM-GNN           &  - &    0.74910    \\ 
MariusGNN           &  - &    0.76950    \\
\midrule
Cluster-GCN (Metis)        & 0.89375  &   0.74553       \\ 
Cluster-GCN (LPMetis)      & \textbf{0.89525}  &  \textbf{0.75272} \\ 
\midrule
Cluster-GAT (Metis)       & 0.8985  &  0.7923 \\ 
Cluster-GAT (LPMetis)     & \textbf{0.8998}  &  \textbf{0.7961} \\ 
\midrule
Cluster-SAGE (Metis)     & 0.9212  &  0.7897 \\ 
Cluster-SAGE (LPMetis)     & \textbf{0.9232}  &  \textbf{0.7926} \\ 
\bottomrule
\end{tabular}
\end{table}
\subsubsection{Evaluation on GNNs with LPMetis}
We examined the efficacy of LPMetis across three different versions of Cluster-GCN, using identical default parameters. 
The results represent the average of five trials. The sole distinction was the replacement of METIS partitioning results (50 sub-graphs).
As shown in Table~\ref{tab:OGB-PARTITION2}, LPMetis surpasses METIS by 0.96\%.
Consequently, LPMetis leads to improvements in the Original, GAT, and SAGE versions of Cluster-GCN, indicating that the LPMetis partitioning algorithm is more congruent with GNNs and comparable performance to METIS when working with small datasets.
Cluster-GAT~\footnote{source code is available at ~\url{https://github.com/dmlc/dgl/tree/master/examples/pytorch/ogb/cluster-gat}} and Cluster-SAGE~\footnote{source code is available at ~\url{https://github.com/dmlc/dgl/tree/master/examples/pytorch/ogb/ogbn-products/graphsage}} are modified versions of Cluster-GCN implemented in OGB. Both approaches utilize Métis for graph partitioning and train subgraph GNN models using GraphSAGE~\cite{hamilton2018inductiverepresentationlearninglarge} and GAT~\cite{veličković2018graphattentionnetworks}.
By employing our LPMetis graph partitioning method and flexibly selecting subgraph GNN models, we can achieve superior node classification performance compared to state-of-the-art large-scale GNN algorithms like MariusGNN and LLM-GNN (which integrates large language models).
\begin{table}[!t]
\centering
\caption{The performance of LPMetis and the SOTA competitors on Dataset $B_2$. LPA and FF algorithms are unsuitable since they generate large sub-graphs that exceed GPU capabilities. FFD produces a large number of small sub-graphs, which is also inappropriate for GNNs.}
\label{tab:PartitionSOTA}
\begin{threeparttable}  
\begin{tabular}{cccc}
\toprule
\textbf{Algorithm}   &  \textbf{Precision} &  \textbf{Recall} &  \textbf{F1} \\ 
\midrule
        \model (RGP)    & 75.29\% & 75.68\% & 75.44\% \\ 
        \model (LPMetis)   & \textbf{76.03\%} & \textbf{76.10\%} & \textbf{76.06\%} \\ 
\bottomrule
\end{tabular}
\end{threeparttable} 
\end{table}
%

For larger datasets, Table \ref{tab:PartitionSOTA} shows that under the LPS-GNN framework on the billion-edge graph $B_2$, LPmetis outperforms other algorithms like LPA (Label Propagation algorithm), FF (Fastunfolding), FFD ( Fastunfolding With Density) and RGP, improving GCN compatibility by 0.55\% to 0.98\%. 
While FFD creates too many small subgraphs and both FF and LPA generate excessively large subgraphs unsuitable for GNN training, LPMetis not only achieves a lower Edge-Cut (EC) but also preserves more valuable edges, enhancing feature smoothing for GCNs. Moreover, LPMetis's performance increases with graph size, retaining more edges and outperforming baselines, despite RGP offering better balance but being less efficient.

\begin{table*}[!t]
\caption{The performance of graph partition.}
\label{tab:my-table}
\resizebox{1\columnwidth}{!}{
\begin{tabular}{ccccccccccccc}
\toprule
\multirow{2}{*}{\textbf{Method}} &
  \multicolumn{4}{c}{\textbf{$\mathbf{A_1}$, K=100}} &
  \multicolumn{4}{c}{\textbf{$\mathbf{A_4}$, K=500}} &
  \multicolumn{4}{c}{\textbf{$\mathbf{B_1}$, K=1000}} \\ 
 &
  \textbf{Time(min)↓} &
  \textbf{STD↓} &
  \textbf{BAL↓} &
  \textbf{Edge-Cut↓} &
  \textbf{Time(min)↓} &
  \textbf{STD↓} &
  \textbf{BAL↓} &
  \textbf{Edge-Cut↓} &
  \textbf{Time(min)↓} &
  \textbf{STD↓} &
  \textbf{BAL↓} &
  \textbf{Edge-Cut↓} \\ 
  \cmidrule(l){2-5}\cmidrule(l){6-9}\cmidrule(l){10-13}
\textbf{FF} &
  92.68 &
  1051.62 &
  11432.12 &
  33.86\% &
  237.05 &
  28447.77 &
  8482.17 &
  52.19\% &
  387.27 &
  101077.39 &
  19195.05 &
  70.14\% \\
\rowcolor{gray!50} \cellcolor{white}
\textbf{FFD} &
  26.62 &
  4.14 &
  2360.83 &
  71.16\% &
  34.67 &
  10.65 &
  269.65 &
  91.33\% &
  OOM &
  OOM &
  OOM &
  OOM \\
\textbf{LPA} &
  5.03 &
  56.64 &
  12283.81 &
  49.50\% &
  94.08 &
  10345.27 &
  120765.23 &
  40.80\% &
  148.82 &
  92405.33 &
  100947.25 &
  55.96\% \\
\rowcolor{gray!50} \cellcolor{white}
\textbf{RGP} &
  33.93 &
  1477.8 &
  \textbf{1.18} &
  45.50\% &
  176.3 &
  4990.48 &
  \textbf{0.23} &
  90.47\% &
  158.55 &
  \textbf{554.29} &
  \textbf{0.0025} &
  99.75\% \\
\textbf{METIS} &
  \textbf{12.23} &
  \textbf{182.53} &
  2.35 &
  \textbf{27.46\%} &
  OOM &
  OOM &
  OOM &
  OOM &
  OOM &
  OOM &
  OOM &
  OOM \\
\rowcolor{gray!50} \cellcolor{white}
\textbf{LPMetis} &
  15.42 &
  656.84 &
  2.22 &
  41.33\% &
  \textbf{43.04} &
  \textbf{2685.33} &
  0.91 &
  \textbf{62.75\%} &
  \textbf{67.22} &
  4584.02 &
  0.28 &
  \textbf{92.76\%} \\ 
\bottomrule
\end{tabular}
}
\end{table*}
\subsubsection{Competitions with SOTA Partition Algorithms}
We evaluated LPMetis' performance on three representative datasets, as detailed in Table~\ref{tab:my-table}. Community detection algorithms, including LPA~\cite{raghavan2007near}, FF~\cite{blondel2008fast}, and FFD~\cite{chen2018network}, excel in speed and edge-cutting efficiency. 
However, these methods do not consistently maintain STD and BAL, whereas METIS and RGP surpass others in these metrics. This necessitated a thorough comparison of our model with these algorithms based on Time, Edge-Cut, STD, and BAL. Thus, we chose these five algorithms as baselines for our research. During experimentation, we examined the properties of these metrics. A low STD/BAL ratio suggests that sub-graph sizes are more uniform, which is conducive to GNN training. 
%
%

\begin{figure}[!t] 
\centering
\includegraphics[width=0.78\linewidth]{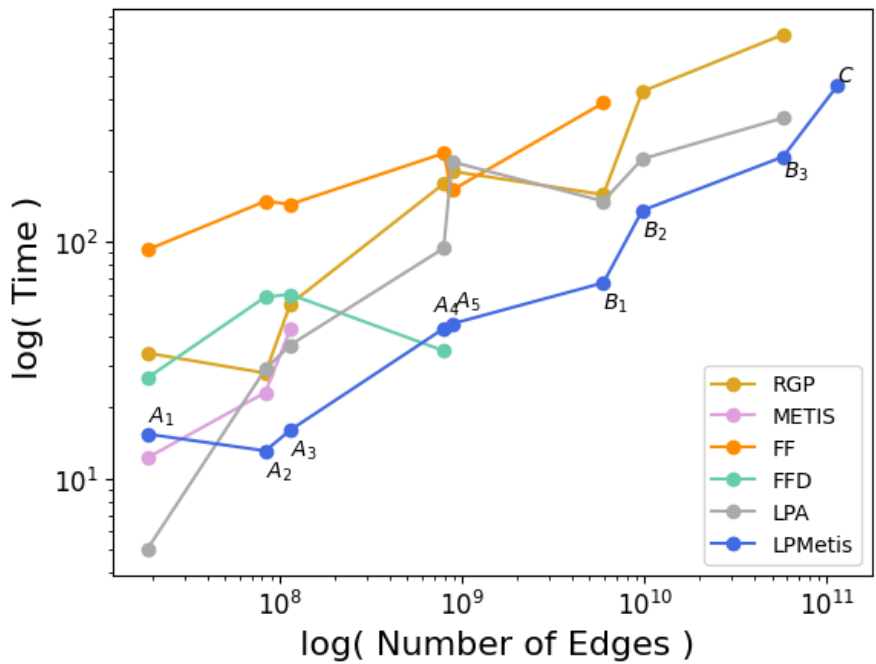} 
\caption{Running time (log scale) of different graph partition methods with reference to the number of edges (log scale). } 
\label{fg:time}
\end{figure}

On the dataset $A_1$, LPMetis's runtime is comparable to that of METIS but is surpassed by LPA (Figure~\ref{fg:time}). However, as the graph size expands, the advantages of both METIS and LPA diminish quickly. 
LPA's dependency on asynchronous node label updating to avoid dichotomy results  in a runtime fluctuation of 231.98\% on datasets of the same magnitude, such as $A_4$ and $A_5$. Furthermore, the formation of large communities leads to load imbalance that affects computational efficiency. 
As a result, LPMetis achieves an average runtime that is only 40.08\% of LPA's.
RGP's time consumption on $A_5$ is 125.06\% of that on $B_1$, even though the scale of nodes and edges in $A_5$ is only $33.22\%$ and $14.78\%$ of $B_1$, respectively. This discrepancy is likely attributable to RGP's initialization mechanism, which initially partitions the graph into sub-graphs randomly, posing challenges in adapting to various graph distributions.
In contrast, LPMetis delegates the acceleration of processing to LPA and the responsibility of sub-graph size adjustment to Metis, endowing LPMetis with consistent robustness and linear scalability across diverse data scales and distributions.

It is critical to acknowledge that the application of RGP results in a significant reduction in STD and BAL for the $B_1$ dataset compared to the $A_4$ dataset. This decline can be primarily attributed to RGP's high Edge-Cut percentage of $99.75\%$ in the $B_1$ dataset. As a consequence, the Maximum Load in BAL becomes inconsequential, rendering BAL an ineffective metric. The STD metric experiences a similar loss of effectiveness.
RGP struggles to process graphs with hundreds of billions of edges within a 24-hour span, whereas the time consumption of LPMetis exhibits near-linear growth. This behavior is attributable to LPMetis's linear computational complexity of $O(|V|+|E|)$ and the rapid convergence of Multi-Level Label Propagation, typically achieving convergence within just 2 to 3 iterations across all tested datasets.
In summary, as the graph size increases, LPMetis achieves SOTA results in these metrics, surpassing other partitioning algorithms.
\subsection{Performance Comparison} \label{secEXPOnlinePerformance}
We begin by comparing the \model with other large-scale GNNs in an offline setting and then present ablation studies of the \model. 
Finally, we show the online performance of \model in three different applications. 
\subsubsection{Offline Comparison with Large-Scale GNN Methods} 
\label{secEXPComparison}
Utilizing the LPS-GNN, both GCN and GraphMAE outperform other algorithms, as illustrated in Table \ref{tab:LS-GNN SOTA}. A 7-layer GCN from Cluster-GCN is employed in this study. Executing GraphSAINT~\cite{zeng2020graphsaint} and PPRGo~\cite{Bojchevski_2020} directly on Dataset $B_2$ results in an Out-of-Memory (OOM) error, given that GraphSAINT sampling requires loading the entire graph into memory, a feat unachievable with extremely large datasets. The Angel Graph~\cite{jiang2020psgraph} provides optimized versions of GAMLP~\cite{2108.10097}, SGC~\cite{wu2019simplifying} and GraphSAGE~\cite{hamilton2017inductive} (Angel GAMLP, Angel SGC and Angel GraphSAGE), enabling them to handle distributed processing of large graphs. 
Angel's training phase employs 10 Parameter Server workers and 70 Spark Cluster executors, consuming 2160G memory. In contrast, \model utilizes a single P40 GPU and only 60 executors, resulting in a 15\% reduction in memory usage and an 18.53\% decrease in the preprocessing time (Table \ref{tab:LS-GNN SOTA}). Though PPRGo is capable of handling billion-scale graphs through an efficient PPR implementation, this would constitute a separate task.
Meticulously designed systems such as ByteGNN~\cite{zheng2022bytegnn}, AliGraph~\cite{zhu2019aligraph}, DistDGLv2~\cite{2112.15345}, and Angel Graph~\cite{jiang2020psgraph} have significant deployment and operation costs on clusters compared to \model.
Within the \model framework, ordinary GNNs can yield results comparable to Angel GAMLP.
%
\begin{table}[!t]
\centering
\caption{The performance of \model and SOTA competitors on Dataset $B_2$. ET and PT denote epoch time and preprocessing time, respectively. }
\label{tab:LS-GNN SOTA}
\begin{tabular}{ccrcc}
\toprule
\textbf{Model}   & \textbf{AUC} & \textbf{\#Parameters} & \textbf{ET} & \textbf{PT}  \\ 
\midrule
Angel GraphSAGE        & 0.591  &   24,320    & 150s  & 178min \\ 
Angel SGC              & 0.806  & 167,386  & \textbf{50s}  & 198min  \\ 
GraphSAINT       &  OOM  & -  & -  & -\\ 
PPRGo            &  OOM  & -  & - & - \\ 
Angel GAMLP            & 0.823  & 502,158  & 1259s &  596min \\ 
\midrule
\model (GCN)      & 0.833  & 123,648  & 73s  & \textbf{145min} \\ 
\model (GraphMAE) & \textbf{0.875} &  \textbf{15,026}   &  71s  & 163min  \\ 
\bottomrule
\end{tabular}
\end{table}
\subsubsection{Ablation on Sampling}  \label{secEXPSampling}
We executed five sampling runs and reported the average results in Table \ref{tab:sampling}. LPMetis partitioned dataset $B_2$ into 6000 sub-graphs. When all sub-graphs were exploited for training, the achieved $F1$ score was merely 0.7693. However, by trimming the sampling ratio to 5\%, we observed a 5x speed increase and a 6.4\% accuracy enhancement. This finding indicates that large graphs frequently contain significant noise and information redundancy. Thus, the optimal sub-graph sampling rate may vary in different scenarios.
Furthermore, we compared various partitions and observed that they predominantly represent sub-graphs from different games (different data domains).
Viewed from another perspective, the "sharing" parameter method employed in sampling is also facilitating cross-domain learning in GNNs, thereby discerning the unique characteristics of social networks across various games. This indirectly suggests that the \model framework may be conducive to the subsequent undertaking of large-scale pretraining for GNNs. However, this ventures into a more profound discourse and warrants a detailed exploration, which we defer for future discussion.
\begin{table}[!t]
\centering
  \caption{ Study with different Sub-graph Sampling Ratio on Dataset $B_2$}
  \label{tab:sampling}
  \begin{tabular}{cccc}
\toprule
\textbf{Method}       & \textbf{Time (hours)} & \textbf{Epochs} & \textbf{F1}  \\
\midrule
    all sub-graphs   &  16h  &  3  & 0.7693    \\
    $10\%$ of the sub-graphs  & 7h  &  20  & 0.8183    \\
    $5\%$ of the sub-graphs  & \textbf{3h} & 20  & \textbf{0.8269}    \\
\bottomrule
\end{tabular}
\end{table}
\begin{table}[!t]
\centering
\caption{F1, Precision, and Recall Score of Sub-Graph Augmentation at Supervised and Unsupervised Tasks on Game Data $B_2$. FA denotes Feature Augmentation. SR denotes Structure Refinement. }
\label{tab:Ablation}
\begin{threeparttable}  
\begin{tabular}{lccc}
\toprule
\textbf{Method}       & \textbf{Precision} & \textbf{Recall} & \textbf{F1-score} \\ 
\midrule
XGBoost                   & 0.7196             & 0.5731          & 0.6380             \\  
\midrule
+\model (wo FA\&SR)  & 0.8512             & 0.8039          & 0.8269            \\ 
+\model (SR-Rand)& 0.8591             & 0.7972          & 0.8270             \\  
+\model (SR-N2V) & 0.8513             & 0.7932          & 0.8212            \\  
+\model (FA)	&  0.8672	& 0.8089	& 0.8370 \\
+\model (SR-PR) & \textbf{0.8972}    & 0.8089 & 0.8508    \\  
+\model (SR-PR+FA) & 0.8925   & \textbf{0.8199} & \textbf{0.8547}    \\  
\bottomrule
\end{tabular}
\end{threeparttable}  
\end{table}
\subsubsection{Ablation on Sub-Graph Augmentation} \label{secEXPsub-graph}
Sub-Graph Augmentation includes both Feature Augmentation and Structure Refinement. Initially, we evaluated the embedding performance of \model without enhancing the features and improving the structure, employing XGBoost~\cite{chen2016xgboost}. As demonstrated in Table \ref{tab:Ablation}, we observed a substantial $29.6\%$ F1 score increase, validating the efficacy of the fundamental \model framework.
Next, we explored Structure Refinement by either randomly removing edges from the sub-graph (SA-Rand) or adding edges based on node2vec similarity (SA-N2V). SA-N2V returned inferior results compared to SA-Rand. Nevertheless, we made further progress by integrating domain knowledge 
through PageRank analysis, denoted as SA-PR. By eliminating the top $5\%$ of edges with the smallest weights, the overall effect was boosted.
Lastly, we fused the global features (coarsened graph embedding) with the sub-graph features. 
Global features improves Recall and F1 score, but to a lesser extent than sub-graph augmentation. This is likely due to the reduced influence of global information in ultra-large graphs, where the individual sub-graphs are considerably large.
In summary, LPMetis achieves SOTA in terms of partitioning time and effectiveness. Feature Augmentation (FA) and Structure Refinement (SR) significantly enhance the performance of GNNs, while Sampling substantially reduces training time without compromising GNN effectiveness. Moreover, \model offers superior flexibility, requiring only minor alterations to the adaptation code for input and output, thus ensuring compatibility with various GNN types.
%
%
\subsubsection{Online Performance} \label{sec:OnlinePerformance}
Tencent Games utilizes three common strategies to improve player retention: acquiring new users through platforms like WeChat and QQ (User Acquisition), employing social recommendations to connect players with game communities (Social Recommendation), and identifying and mitigating fraudulent users that disrupt the game ecosystem (Identification of Fraudulent Users).
But online systems maintain strict controls over experimental traffic to protect business performance and follow meticulous deployment processes. These factors limit the scale of experiments that can be conducted. To incorporate embeddings as features in online experiments, we utilize DGI. This approach facilitates seamless integration into the online environment while enabling objective effectiveness comparisons by changing only the feature variables. Additionally, SOTA algorithms utilized in online scenarios have demonstrated their robustness over time, even if they may seem slightly outdated.
We observed an 11.94\% improvement in the fraudulent user detection rate over PU-DNN, which directly combines Positive and Unlabeled Learning with DNN, as depicted in Table \ref{tab:onlineperformance}. LPS-GNN also outperformed the feature propagation-based GNN algorithm FS-GNN by a large margin. Furthermore, we achieved a 13.89\% improvement over DeepFM (a widely used recommendation algorithm) in advertising for user acquisition scenarios (100-billion graph).  
we assessed the performance of \model with GCN in the friend recommendation task and observed an 8.24\% increase in conversion rate compared to the SOTA method XGB. 
In various online tasks, including supervised and unsupervised settings, \model consistently achieves SOTA results.
\hide{
\begin{table}[t]
\caption{Online performance of three different applications in Tencent. TT denotes Total Training Time. SR stands for Social Recommendation, IFU denotes Identification of Fraudulent Users and UA refers to User Acquisition within online games.}
\centering
\label{tab:onlineperformance}
\footnotesize
\resizebox{1\columnwidth}{!}{%
\begin{tabular}{ccclll}
\toprule
\textbf{Type} & \textbf{Scenario} & \textbf{DataSet} & \textbf{Algorithm} & \textbf{Precision} & \textbf{TT} \\
\midrule
\multirow{2}{*}{Supervised} & \multirow{2}{*}{SR} & \multirow{2}{*}{$A_4$} & XGBoost & 24.86\% & - \\
 &  &  & \model (GCN) & \textbf{26.91\%}(+8.24\%) & 2.71h \\
\midrule
\multirow{4}{*}{Unsupervised} & \multirow{2}{*}{IFU} & \multirow{2}{*}{$B_2$} & PU-DNN & 19.50\% & - \\
 &  &  & \model (DGI) & \textbf{21.83\%}(+11.94\%) & 4.29h \\
 \cmidrule{2-6}
 & \multirow{2}{*}{UA} & \multirow{2}{*}{$C$} & DeepFM & 0.36\% & - \\
 &  &  & \model (DGI) & \textbf{0.41\%}(+13.89\%) & 9.73h \\
 \bottomrule
\end{tabular}
}
\end{table}
}
\begin{table}[!t]
\caption{Online performance of three different applications in Tencent. TT denotes Total Training Time. SR stands for Social Recommendation, IFU denotes Identification of Fraudulent Users and UA refers to User Acquisition within online games.}
\centering
\label{tab:onlineperformance}
\footnotesize
\begin{tabular}{ccclll}
\toprule
\textbf{Type} & \textbf{Scenario} & \textbf{DataSet} & \textbf{Algorithm} & \textbf{Precision} & \textbf{TT} \\
\midrule
\multirow{2}{*}{Supervised} & \multirow{2}{*}{SR} & \multirow{2}{*}{$A_4$} & XGBoost & 24.86\% & - \\
 &  &  & \model (GCN) & \textbf{26.91\%}(+8.24\%) & 2.71h \\
\midrule
\multirow{5}{*}{Unsupervised} & \multirow{3}{*}{IFU} & \multirow{3}{*}{$B_2$} & PU-DNN & 19.50\%  & - \\
 &  &  & FS-GNN & 20.93\% (+7.33\%) & - \\
 &  &  & \model (DGI) & \textbf{21.83\%}(+11.94\%) & 4.29h \\
 \cmidrule{2-6}
 & \multirow{2}{*}{UA} & \multirow{2}{*}{$C$} & DeepFM & 0.36\% & - \\
 &  &  & \model (DGI) & \textbf{0.41\%}(+13.89\%) & 9.73h \\
 \bottomrule
\end{tabular}
\end{table} 
\section{Conclusion}
In conclusion, this paper introduces the \model framework, intended to boost performance and expedite computations of GNNs on super-large-scale graphs comprising over 100 billion edges. This versatile framework facilitates the seamless integration of various SOTA GNN algorithms and supports multiple downstream settings, including supervised, unsupervised, and semi-supervised approaches. The effectiveness of the proposed \model has been verified in diverse scenarios, resulting in significant improvements. Notably, the \model has been successfully implemented on the Tencent platform and applied to several real-world scenarios within Tencent Games. To expand the framework's capabilities, we aim to explore the following promising avenues in future research. Firstly, we plan to enhance the \model to manage heterogeneous networks, thereby accommodating a broader spectrum of real-world scenarios. Secondly, we intend to examine the application of the \model to dynamic graphs, characterized by the ongoing addition or removal of nodes and edges. Lastly, we will verify the applicability of the \model to graph transfer learning.

\bibliographystyle{ACM-Reference-Format}
\bibliography{sample-base}

\section{Appendices}
%
%
%
\subsection{Implementation Details} \label{secEXPImplementationDetails}
For the software versions, we installed Spark 2.3, TensorFlow 1.12.0, and PyTorch 1.9.0. METIS 5.1.0 was obtained from Website \footnote{\url{http://glaros.dtc.umn.edu/gkhome/metis/metis/download}} and a Python wrapper \footnote{\url{https://metis.readthedocs.io/en/latest/}} for the METIS library.
The implementation of \model is carried out using the distributed computing infrastructure provided by Tencent.
GNN Preprocessing Stage (such as decoupling base algorithms) and Graph Partition were conducted using several distributed computing platforms within the company, including the parameter server (PS) platform, a scheduling platform capable of executing SPARK tasks, and a distributed file system. In Section~\ref{secEXPComparison}, we provide additional runtime information on dataset $B_2$. During the runtime of the Angel Parameter Server, both Spark and PS resources are required. One Spark executor occupies approximately 2 CPU resources and 30GB of memory, while one PS worker uses 2 CPUs and 6GB of memory. The driver consumes fewer resources, which we have simplified and disregarded for the sake of brevity. The training of GNNs was conducted on a Linux server equipped with 253GB of RAM, a P40 GPU, and a 48-core AMD CPU. 
Using Table \ref{tab:exp_setting}, we can observe that for LPS-GNN, by utilizing an additional GPU and reducing memory usage by $15\%$, the preprocessing time decreases from 178 minutes to 145 minutes, the efficiency improves by at least $18.53\%$. Remarkably, the flexibility of the framework remains largely unaffected. 
\begin{table}[!t]
\label{tab:exp_setting}
\caption{Experiment Settings on DataSet $B_2$}
\begin{tabular}{cccccc}
\toprule
\textbf{Model} & \textbf{Parameter Servers} & \textbf{Spark} & \textbf{CPUs} & \textbf{GPU} & \textbf{Memory} \\ 
\midrule
Angel GraphSAGE & 10 workers & 70 executors & 160 & 0 & 2160G \\
\rowcolor{gray!50} \cellcolor{white}
Angel SGC & 10 workers & 70 executors & 160 & 0 & 2160G \\ 
Ange GAMLP & 10 workers & 70 executors & 160 & 0 & 2160G \\ 
\rowcolor{gray!50} \cellcolor{white}
\model (GCN) & 0 workers & 60 executors & 120 & 1(P40) & 1824G\\
\model (GraphMAE) & 0 workers & 60 executors & 120 & 1(A100) & 1840G \\
\bottomrule
\end{tabular}
\end{table}
\subsection{LPMetis}
\label{appendix:tab-my-table2}
We conducted comparative experiments to evaluate the effectiveness of LPMetis on six different data sizes. 
Table \ref{tab:my-table} provides the performance of LPMetis on three representative datasets, while Table \ref{tab:my-table2} presents the remaining three datasets, showing similar overall numerical distributions.
%
\begin{table*}[!t]
\caption{The performance of graph partition.}
\label{tab:my-table2}
\resizebox{1\columnwidth}{!}{
\begin{tabular}{ccccccccccccc}
\toprule
\multirow{2}{*}{\textbf{Method}} &
  \multicolumn{4}{c}{\textbf{$\mathbf{A_2}$, K=100}} &
  \multicolumn{4}{c}{\textbf{$\mathbf{A_3}$, K=100}} &
  \multicolumn{4}{c}{\textbf{$\mathbf{A_5}$, K=500}} \\ 
 &
  \textbf{Time(min)↓} &
  \textbf{STD↓} &
  \textbf{BAL↓} &
  \textbf{Edge-Cut↓} &
  \textbf{Time(min)↓} &
  \textbf{STD↓} &
  \textbf{BAL↓} &
  \textbf{Edge-Cut↓} &
  \textbf{Time(min)↓} &
  \textbf{STD↓} &
  \textbf{BAL↓} &
  \textbf{Edge-Cut↓} \\ 
  \cmidrule(l){2-5}\cmidrule(l){6-9}\cmidrule(l){10-13}
\textbf{FF} &
  148.75 &
  8497.54 &
  1446.48 &
  15.08\% &
  144.38 &
  13288.19 &
  37188.64 &
  33.86\% &
  165.58 &
  14065.86 &
  4899.86 &
  52.80\% \\
\rowcolor{gray!50} \cellcolor{white}
\textbf{FFD} &
  58.6 &
  27.08 &
  140.22 &
  75.91\% &
  60.02 &
  6.26 &
  1764.22 &
  87.71\% &
  OOM &
  OOM &
  OOM &
  OOM \\
\textbf{LPA} &
  29 &
  59.93 &
  272.66 &
  33.82\% &
  36.7 &
  20857.07 &
  233184.59 &
  5.03\% &
  218.25 &
  34382.25 &
  706049.28 &
  14.14\% \\
\rowcolor{gray!50} \cellcolor{white}
\textbf{RGP} &
  27.92 &
  13950.79 &
  \textbf{0.59} &
  61.54\% &
  54.68 &
  5134.22 &
  \textbf{0.67} &
  81.29\% &
  199.15 &
  5097.07 &
  \textbf{0.19} &
  88.64\% \\
\textbf{METIS} &
  23.12 &
  \textbf{7036.53} &
  1.14 &
  \textbf{13.73\%} &
  42.83 &
  \textbf{3153.17} &
  2.34 &
  \textbf{42.91\%} &
  OOM &
  OOM &
  OOM &
  OOM \\
\rowcolor{gray!50} \cellcolor{white}
\textbf{LPMetis} &
  \textbf{13.1} &
  8852.83 &
  0.83 &
  39.37\% &
  \textbf{16.03} &
  3432.4 &
  1.63 &
  63.65\% &
  \textbf{45.1} &
  \textbf{3069.91} &
  0.69 &
  \textbf{60.18\%} \\ 
\bottomrule
\end{tabular}
}
\end{table*}
\subsubsection{Different Types of Coarsening} \label{secDifferentTypesofCoarsening}
The ``Coarsening by Edge'' technique maintains the count of nodes from the antecedent coarsened graph, $\tilde{G}_{t-1}$, as the value for the extant node $\tilde{v}_i$, as explicitly demonstrated in Algorithm \ref{alg:Coarsen} (Lines 2 to 4). Conversely, within the ``Coarsening by Node'' paradigm, the value ascribed to node $v_i$ equates to the aggregate number of nodes present in the primordial graph (Lines 6 to 9).
For an illustrative exemplar, one might refer to the extreme right of Figure \ref{fig:lpmetis}, wherein ``Different Types of Coarse'' visually explicates a simplistic instance. Within this scenario, the ``by Node'' method computes the valuation of the yellow nodes in $\tilde{G_2}$ by totaling the values of all corresponding yellow nodes in $\tilde{G_1}$, yielding $26=5+5+4+7+5$. 
In contrast, the ``by Edge'' strategy solely acknowledges the number of yellow nodes extant in $\tilde{G_1}$, represented by $\tilde{v}_{yellow}=5$.
Pertaining to the computation of edge weights, there exists a uniformity in methodology across both strategies; this involves the summation of the edges interconnecting partitions. To illustrate, the edge weight of $\tilde{G_2}$ interfacing the yellow and blue nodes, denoted as $e=0.3+0.4=0.7$, is derived from the cumulative sum of all proximate yellow and blue edges in $\tilde{G_1}$. For additional specificity regarding this process, Algorithm \ref{alg:Coarsen} (Lines 11 to 16) is a comprehensive reference.k%
\begin{algorithm}[!t]
    \caption{ Coarsening Graph by Node or Edge (Coarsen) }
    \label{alg:Coarsen}
    \begin{algorithmic}[1]
        \REQUIRE {$\{g_1, g_2,\ldots,g_M\}, Type, G=(V, E)$}
        \ENSURE  {Coarsened Graph $\tilde{G^{\prime}}=(\tilde{V^{\prime}}, \tilde{E^{\prime})}$ }

        \IF{ Type == ``Edge'' }
            \FOR {$g_i$ in $\{g_1, g_2,\ldots,g_M\}$}
                \STATE $\tilde{V^{\prime}_i} \gets | g_i| $, where  $|g_i|$ is the number of nodes in the sub-graph $g_i$
            \ENDFOR
        \ELSE 
            \FOR {$v_{j}$ in $V$}
               \STATE $i \gets SubGraphID(v_{j}) $ Find the ID of the subgraph to which $v_j$ belongs.
               \STATE  $ \tilde{V^{\prime}_i} \gets \tilde{V^{\prime}_i} + v_j$
            \ENDFOR
            
        \ENDIF

        \STATE $\tilde{E^{\prime}} \gets 0$ Init the value of all edges to zero. 

        \FOR {$e_{ij}$ in $E$}
            \STATE $m \gets SubGraphID(v_i)$ 
            \STATE $n \gets SubGraphID(v_j)$
            \STATE  $\tilde{E^{\prime}_{mn}} \gets \tilde{E^{\prime}_{mn}} + e_{ij} $
        \ENDFOR
       \ENSURE  {$ \tilde{G^{\prime}}  $}
    \end{algorithmic}
\end{algorithm}
%
\subsection{Offline Comparison Experiments of \model} 
We provide additional runtime information on dataset $B_2$ for GNN offline comparison experiments. During the runtime of Angel Parameter Server, both Spark and PS resources are required. One Spark executor occupies approximately 2 CPU resources and 30GB of memory, while one PS worker uses 2 CPUs and 6GB of memory. The driver consumes fewer resources, which we have simplified and disregarded for the sake of brevity. 
\subsection{ Proofs of Theorem \ref{theorem:1}}
Applying the Lipschitz property of $ \| L \| \leq 1$ and further decompose the internal term, yielding:
\begin{align*}
\|H_{g_i}^{(k+1)} - H_{g_i}^{(k)}\| &\leq \| L \|\|\tilde{A}_{g_i} H_{g_i}^{(k)} W^{(k+1)} - \tilde{A}_{g_i} H_{g_i}^{(k-1)} W^{(k)}\|  \\
&\leq \| L \|\|\tilde{A}_{g_i}\| \left( \|H_{g_i}^{(k)} \Delta W^{(k)}\| + \|\Delta H_{g_i}^{(k)} W^{(k)}\| \right)
\end{align*}
where $\Delta W^{(k)} = W^{(k+1)} - W^{(k)}$, $\Delta H^{(k)} = H^{(k)} - H^{(k-1)}$.
 
a) Let $\tilde{A} = D^{-1/2} (A + I) D^{-1/2}$ be the renormalized adjacency matrix. The eigenvalue range of $\tilde{A}$ is $0 < \lambda_i \leq 1$, and the spectral norm $\|\tilde{A}\| \leq 1$. By Lemma \ref{lemma:Courant-Fischer}, we have $\|\tilde{A}_{g_i}\| \leq \| \tilde{A} \| \leq 1$.
b) When the GNN algorithm guarantees convergence, assuming $\|W^{(k)}\| \leq 1$,  $\|H_{g_i}^{(k)}\| \leq C$, then $\|W^{(k+1)} - W^{(k)}\| \leq \epsilon \ll 1$, we have $\|H_{g_i}^{(k)} \Delta W^{(k)}\| \leq C \cdot \epsilon $, where $C \epsilon$ can be ignored. 
Thus, $\|H_{g_i}^{(k+1)} - H_{g_i}^{(k)}\| \leq \| L \|\|\tilde{A}_{g_i}\|\|H_{g_i}^{(k)} - H_{g_i}^{(k-1)}\| \|W^{(k)}\|$. Since $\| L \|\|\tilde{A}_{g_i}\| \|W^{(k)}\|  \leq 1$, according to the Banach fixed-point theorem, we have $\|H_{g_i}^{(k+1)} - H_{g_i}^{(k)}\| \leq \|H_{g_i}^{(k)} - H_{g_i}^{(k-1)}\|$. Therefore, the iterative process converges.
%
%
%
\begin{lemma} \label{lemma:Courant-Fischer}
The spectral norm $\|G\|$ of a matrix $G$ is greater than or equal to the spectral norm $\|g\|$ of any submatrix $g$ within $G$, denoted as $\|G\| \geq \|g\|$.

Proof: $\|G\|$ is its largest absolute eigenvalue.
Let $ g $ be an $ m \times m $ submatrix of $ G $. Construct a projection matrix $ P \in \mathbb{R}^{m \times n} $ where the random $ m $ rows are standard basis vectors and the rest are zeros. We have submatrix $ g = P^\top G P $.
Becuase the spectral norm of $ P $ is $ \|P\| = 1 $, as $\|Px\| \leq \|x\| $ for any vector $x$.
Using the properties of operator norms, $ \|P\| = \|P^\top\| = 1 $ we have: $ \|g\| = \|P^\top GP\| \leq \|P\| \cdot \|G\| \cdot \|P^\top\| \leq \|G\|$.
\end{lemma}

\end{document}